%% file: emnlp2022.tex
\pdfoutput=1

\documentclass[11pt]{article}

\usepackage[]{EMNLP2022}

\usepackage{times}
\usepackage{latexsym}

\usepackage{amsmath}
\usepackage{amssymb}
\usepackage{bm}
\usepackage{booktabs}
\usepackage{graphicx}
\usepackage{multicol}
\usepackage{multirow}
\usepackage{blindtext}
\usepackage{dirtree}

\usepackage[T1]{fontenc}

\usepackage[utf8]{inputenc}
\usepackage{pmboxdraw}

\usepackage{microtype}

\usepackage{inconsolata}

\newcommand{\parheader}[1]{{\bf \smallskip \noindent #1.}}

\newcommand{\DONE}[1]{\noindent \textcolor{green}{\textbf{DONE}}\\ }

\title{What the DAAM: Interpreting Stable Diffusion Using Cross Attention}

\author{Raphael Tang,$^{*1}$ Linqing Liu,\thanks{~~Equal contribution.}$~\hspace{0.5mm}^2$ Akshat Pandey,$^1$ Zhiying Jiang,$^3$ Gefei Yang,$^1$\\ \textbf{Karun Kumar,$^1$ Pontus Stenetorp,$^2$ Jimmy Lin,$^3$ Ferhan Ture$^1$} \vspace{1mm}\\
$^1$Comcast Applied AI\hspace{5mm}$^2$University College London\hspace{5mm}$^3$University of Waterloo\\
{\small $^1$\texttt{{\{raphael\_tang,akshat\_pandey,gefei\_yang,karun\_kumar,ferhan\_ture\}}@comcast.com}} \\
{\small $^2$\texttt{\{linqing.liu,p.stenetorp\}@cs.ucl.ac.uk}}\hspace{3mm}
{\small $^3$\texttt{\{zhiying.jiang,jimmylin\}@uwaterloo.ca}}}

\begin{document}
\maketitle
\begin{abstract}\vspace{-2mm}
Large-scale diffusion neural networks represent a substantial milestone in text-to-image generation, but they remain poorly understood, lacking interpretability analyses.
In this paper, we perform a text--image attribution analysis on Stable Diffusion, a recently open-sourced model.
To produce pixel-level attribution maps, we upscale and aggregate cross-attention word--pixel scores in the denoising subnetwork, naming our method DAAM.
We evaluate its correctness by testing its semantic segmentation ability on nouns, as well as its generalized attribution quality on all parts of speech, rated by humans.
We then apply DAAM to study the role of syntax in the pixel space, characterizing head--dependent heat map interaction patterns for ten common dependency relations.
Finally, we study several semantic phenomena using DAAM, with a focus on feature entanglement, where we find that cohyponyms worsen generation quality and descriptive adjectives attend too broadly.
To our knowledge, we are the first to interpret large diffusion models from a visuolinguistic perspective, which enables future lines of research.
Our code is at \url{https://github.com/castorini/daam}.

\end{abstract}

\section{Introduction}\vspace{-1mm}

Diffusion neural networks trained on billions of image--caption pairs represent the state of the art in text-to-image generation~\cite{yang2022diffusion},
with some achieving realism comparable to photographs in human evaluation, such as Google's Imagen~\cite{saharia2022photorealistic} and OpenAI's DALL-E 2~\cite{ramesh2022hierarchical}.
However, despite their quality and popularity, the dynamics of their image synthesis remain undercharacterized.
Citing ethical concerns, these organizations have restricted the general public from using the models and their weights, preventing effective white-box~(or even blackbox) analysis.
To overcome this barrier, Stability AI recently open-sourced Stable Diffusion~\cite{rombach2022high}, a 1.1 billion-parameter latent diffusion model pretrained and fine-tuned on the LAION 5-billion image dataset~\cite{schuhmann2022laion}.

\begin{figure}
    \centering
    \includegraphics[scale=0.125]{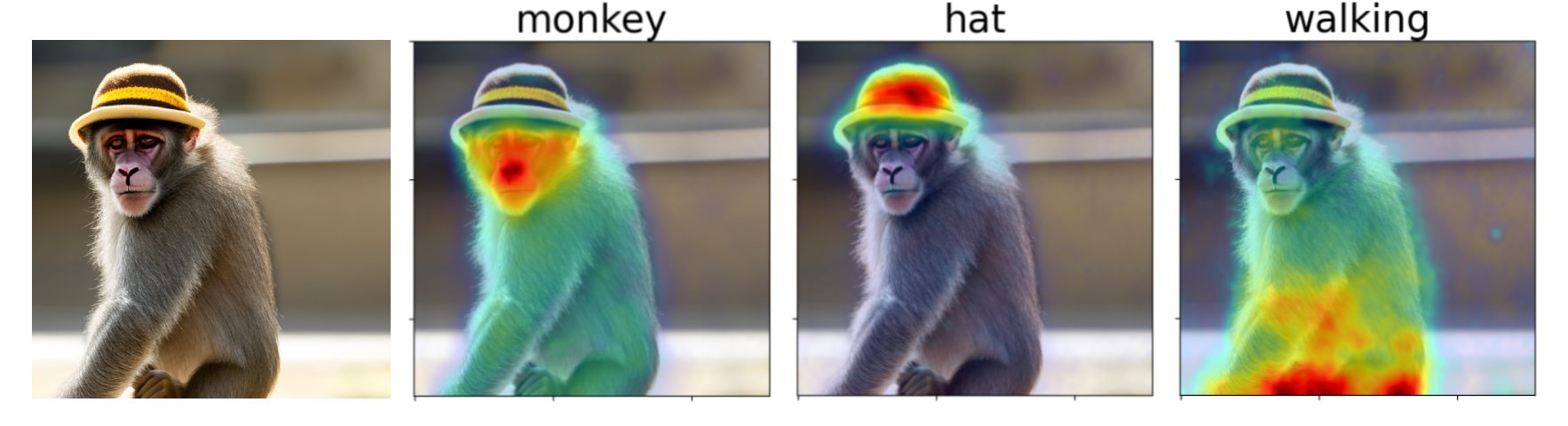}
    \caption{The original synthesized image and three DAAM maps for ``monkey,'' ``hat,'' and ``walking,'' from the prompt, ``monkey with hat walking.''}
    \label{fig:example}
\end{figure}

We probe Stable Diffusion to provide insight into the workings of large diffusion models.
With a focus on text-to-image attribution, our central research question is, \textit{``How does an input word influence parts of a generated image?''}
To this, we first propose to produce two-dimensional attribution maps for each word by combining cross-attention maps in the model, as delineated in Section~\ref{sec:daam-approach}.
A related work in prompt-guided editing from \citet{hertz2022prompt} conjectures that per-head cross attention relates words to areas in Imagen-generated images, but they fall short of constructing global per-word attribution maps.
We name our method diffusion attentive attribution maps, or DAAM for short---see Figure~\ref{fig:example} for an example.

To evaluate the veracity of DAAM, we apply it to a semantic segmentation task~\cite{lin2014microsoft} on generated imagery, comparing DAAM maps with annotated segments.
We attain a 58.9--64.8 mean intersection over union (mIoU) score, which is competitive with unsupervised segmentation models, described in Section~\ref{sec:obj-attr}.
We further bolster these noun attribution results using a generalized study covering all parts of speech, such as adjectives and verbs.
Through human annotation, we show that the mean opinion score~(MOS) is above fair to good~(3.4--4.2) on interpretable words.

Next, we characterize how relationships in the syntactic space of prompts relate to those in the pixel space of images.
We assess head--dependent DAAM map interactions across ten common syntactic relationships, finding that, for some, the heat map of the dependent strongly subsumes that of the head, while the opposite is true for others.
For still others, such as coreferent word pairs, the words' maps greatly overlap, indicating identity.
We assign visual intuition to our observations; for example, we conjecture that the maps of verbs contain those of their subjects, because verbs often contextualize both the subjects and their surroundings.

Finally, we form hypotheses to further examine our syntactic findings, studying semantic phenomena through the lens of DAAM, particularly those affecting the generation quality.
In Section~\ref{sec:cohypo-entanglement}, we demonstrate that, in constructed prompts with two distinct nouns, cohyponyms have worse quality, e.g., ``a giraffe and a zebra'' generates either a giraffe \textit{or} a zebra, but not both.
We observe that cohyponym status and generation incorrectness each increases the amount of overlap between the heat maps.
We also show in Section~\ref{sec:adj-entanglement} that descriptive adjectives attend too broadly across the image, far beyond the nouns they modify.
If we hold the scene layout fixed~\cite{hertz2022prompt} and vary only the adjective, the entire image changes, not just the noun.
These two phenomena suggest feature entanglement, where objects are entangled with both the scene and other objects.

In summary, our contributions are as follows: \textbf{(1)} we propose and evaluate an attribution method, novel within the context of interpreting diffusion models, measuring which parts of the generated image the words influence most; \textbf{(2)} we provide new insight into how syntactic relationships map to generated pixels, finding evidence for directional imbalance in head--dependent DAAM map overlap, alongside visual intuition~(and counterintuition) in the behaviors of nominals, modifiers, and function words; and \textbf{(3)} we shine light on failure cases in diffusion models, showing that descriptive adjectival modifiers and cohyponyms result in entangled features and DAAM maps.

\section{Our Approach}

\subsection{Preliminaries}
Latent diffusion models~\cite{rombach2022high} are a class of denoising generative models that are trained to synthesize high-fidelity images from random noise through a gradual denoising process, optionally conditioned on text.
They generally comprise three components: a deep language model like CLIP~\cite{radford2021learning} for producing word embeddings; a variational autoencoder~(VAE; \citealp{kingma2013auto}) which encodes and decodes latent vectors for images; and a time-conditional U-Net~\cite{ronneberger2015u} for gradually denoising latent vectors.
To generate an image, we initialize the latent vectors to random noise, feed in a text prompt, then iteratively denoise the latent vectors with the U-Net and decode the final vector into an image with the VAE.

Formally, given an image, the VAE encodes it as a latent vector $\bm \ell_{t_0} \in \mathbb{R}^d$.
Define a forward ``noise injecting'' Markov chain
$p(\bm \ell_{t_i} | \bm \ell_{t_{i-1}}) := \mathcal{N}(\bm \ell_{t_i}; \sqrt{1 - \alpha_{t_i}} \bm \ell_{t_0}, \alpha_{t_i} \bm I)$ where $\{\alpha_{t_i}\}_{i=1}^T$ is defined following a schedule so that $p(\bm \ell_{t_T})$ is approximately zero-mean isotropic.
The corresponding denoising reverse chain is then parameterized as 
\begin{equation}\vspace{-2mm}
\small
\resizebox{200pt}{!}{%
    $p(\bm \ell_{t_{i-1}} | \bm \ell_{t_i}) := \mathcal{N}(\bm \ell_{t_{i-1}}; \frac{1}{\sqrt{1 - \alpha_{t_i}}}(\bm \ell_{t_i} + \alpha_{t_i}\epsilon_\theta(\bm \ell_{t_i}, t_i)), \alpha_{t_i}\bm I),$
}
\end{equation}
for some denoising neural network $\epsilon_\theta(\bm \ell, t)$ with parameters $\theta$.
Intuitively, the forward process iteratively adds noise to some signal at a fixed rate, while the reverse process, equipped with a neural network, removes noise until recovering the signal.
To train the network, given caption--image pairs, we optimize
\begin{equation}\vspace{-2mm}
\small
\resizebox{200pt}{!}{%
    $\min_\theta \sum_{i=1}^T \zeta_i\mathbb{E}_{p(\bm \ell_{t_i} | \bm \ell_{t_0})}\lVert \epsilon_\theta(\bm \ell_{t_i}, t_i) - \nabla_{\bm \ell_{t_i}} \log p(\bm \ell_{t_i} | \bm \ell_{t_0}) \rVert^2_2$,
}
\end{equation}
where $\{\zeta_i\}_{i=1}^T$ are constants computed as $\zeta_i := 1 - \prod_{j=1}^i(1 - \alpha_j)$.
The objective is a reweighted form of the evidence lower bound for score matching~\cite{song2021scorebased}.
To generate a latent vector, we initialize $\hat{\bm \ell}_{t_T}$ as Gaussian noise and iterate
\begin{equation}\label{eqn:inference-steps}\vspace{-2mm}
\small
\resizebox{200pt}{!}{%
$\hat{\bm \ell}_{t_{i-1}} = \frac{1}{\sqrt{1 - \alpha_{t_i}}}(\hat{\bm \ell}_{t_i} + \alpha_{t_i}\epsilon_\theta(\hat{\bm \ell}_{t_i}, t_i)) + \sqrt{\alpha_{t_i}}z_{t_i}$.
}
\end{equation}
In practice, we apply various optimizations to improve the convergence of the above step, like modeling the reverse process as an ODE~\cite{song2021scorebased}, but this definition suffices for us.
We can additionally condition the latent vectors on text and pass word embeddings $\bm X := [\bm x_1; \cdots; \bm x_{l_W}]$ to $\epsilon_\theta(\bm \ell, t; \bm X)$.
Finally, the VAE decodes the denoised latent $\hat{\bm \ell}_{t_0}$ to an image.
For this paper, we use the publicly available weights of the state-of-the-art, 1.1 billion-parameter Stable Diffusion 2.0 model~\cite{rombach2022high}, trained on 5 billion caption--image pairs~\cite{schuhmann2022laion} and implemented in HuggingFace's Diffusers library~\cite{platen2022diffusers}.

\subsection{Diffusion Attentive Attribution Maps}\label{sec:daam-approach}
Given a large-scale latent diffusion model for text-to-image synthesis, which parts of an image does each word influence most?
One way to achieve this would be attribution approaches, which are mainly perturbation- and gradient-based~\cite{alvarez2018robustness, selvaraju2017grad}, where saliency maps are constructed either from the first derivative of the output with respect to the input, or from input perturbation to see how the output changes.
Unfortunately, gradient methods prove intractable due to needing a backpropagation pass \textit{for every pixel for all $T$ time steps}, and even minor perturbations result in  significantly different images in our pilot experiments.

Instead, we use ideas from natural language processing, where word attention was found to indicate lexical attribution~\cite{clark2019does}, as well as the spatial layout of Imagen's images~\cite{hertz2022prompt}.
In diffusion models, attention mechanisms cross-contextualize text embeddings with coordinate-aware latent representations~\cite{rombach2022high} of the image, outputting scores for each token--image patch pair.
Attention scores lend themselves readily to interpretation since they are already normalized in $[0, 1]$.
Thus, for pixelwise attribution, we propose to aggregate these scores over the spatiotemporal dimensions and interpolate them across the image.

We turn our attention to the denoising network $\epsilon_\theta(\bm \ell, t; \bm X)$ responsible for the synthesis.
While the subnetwork can take any form, U-Nets remain the popular choice~\cite{ronneberger2015u} due to their strong image segmentation ability.
They consist of a series of downsampling convolutional blocks, each of which preserves some local context, followed by upsampling deconvolutional blocks, which restore the original input size to the output.
Specifically, given a 2D latent $\bm \ell_t \in \mathbb{R}^{w\times h}$, the downsampling blocks output a series of vectors $\{\bm h^{\downarrow}_{i,t}\}_{i=1}^K$, where $\bm h^{\downarrow}_{i,t} \in \mathbb{R}^{\lceil\frac{w}{c^{i}}\rceil \times \lceil\frac{h}{c^{i}}\rceil}$ for some $c > 1$.
The upsampling blocks then iteratively upscale $\bm h^{\downarrow}_{K,t}$ to $\{\bm h^{\uparrow}_{i,t}\}_{i=K-1}^0 \in \mathbb{R}^{\lceil\frac{w}{c^{i}}\rceil \times \lceil\frac{h}{c^{i}}\rceil}$.
To condition these representations on word embeddings, \citet{rombach2022high} use multi-headed cross-attention layers~\cite{vaswani2017attention}
\begin{equation}\small\vspace{-3mm}
    \bm h^{\downarrow}_{i,t} := F^{(i)}_t(\hat{\bm h}^{\downarrow}_{i,t}, \bm X) \cdot (W^{(i)}_v\bm X ),
\end{equation}\vspace{-6mm}
\begin{equation}\small\label{eqn:attn-maps}\vspace{-3mm}
\resizebox{200pt}{!}{%
    $F^{(i)}_t(\hat{\bm h}^{\downarrow}_{i,t}, \bm X) := \text{softmax}\left((W_q^{(i)} \hat{\bm h}^{\downarrow}_{i,t})(W^{(i)}_k \bm X)^T/\sqrt{d}\right)$,
}
\end{equation}
where $F^{(i)\downarrow}_t \in \mathbb{R}^{\lceil\frac{w}{c^{i}}\rceil \times \lceil\frac{h}{c^{i}}\rceil \times l_H \times l_W}$ and $W_k$, $W_q$, and $W_v$ are projection matrices with $l_H$ attention heads.
The same mechanism applies when upsampling $\bm h_i^\uparrow$.
For brevity, we denote the respective attention score arrays as $F^{(i)\downarrow}_t$ and $F^{(i)\uparrow}_t$, and we implicitly broadcast matrix multiplications as per NumPy convention~\cite{harris2020array}.
\begin{figure}
    \centering
    \includegraphics[scale=0.22]{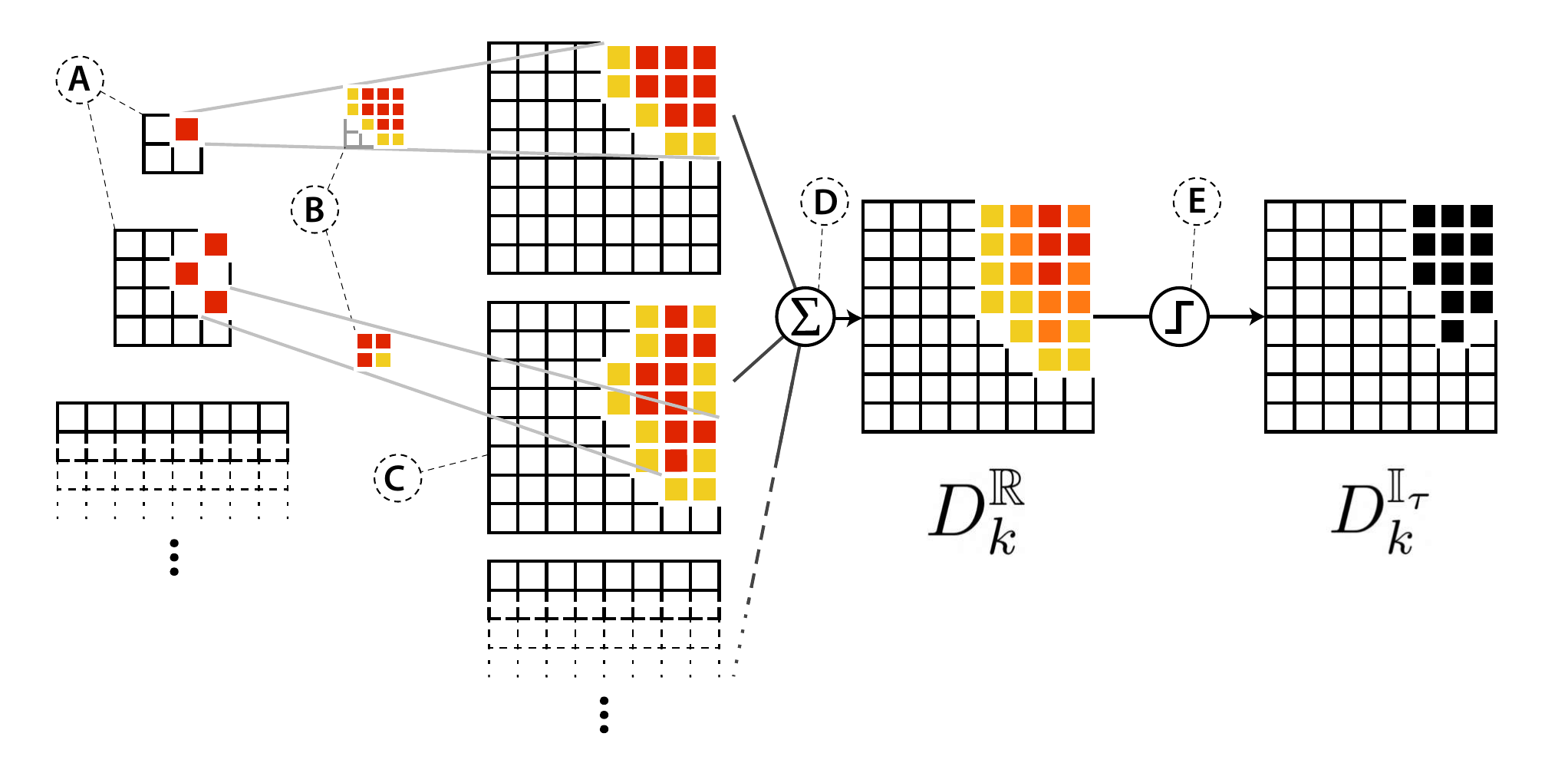}
    \caption{Illustration of computing DAAM for some word: the multiscale attention arrays from Eqn.~\eqref{eqn:attn-maps}~(see \textbf{A}); the bicubic interpolation (\textbf{B}) resulting in expanded maps (\textbf{C}); summing the heat maps across the layers~(\textbf{D}), as in Eqn.~\eqref{eqn:summation}; and the thresholding (\textbf{E}) from Eqn.~\eqref{eqn:threshold}.}
    \label{fig:daam}
\end{figure}

\parheader{Spatiotemporal aggregation}
$F^{(i)\downarrow}_t[x, y, \ell, k]$ is normalized to $[0, 1]$ and connects the $k^\text{th}$ word to the intermediate coordinate $(x, y)$ for the $i^\text{th}$ downsampling block and $\ell^\text{th}$ head.
Due to the fully convolutional nature of U-Net (and the VAE), the intermediate coordinates locally map to a surrounding affected square area in the final image, the scores thus relating each word to that image patch.
However, different layers produce heat maps with varying scales, deepest ones being the coarsest (e.g., $\bm h^\downarrow_{K,t}$ and $\bm h^\uparrow_{K-1,t}$), requiring spatial normalization to create a single heat map.
To do this, we upscale all intermediate attention score arrays to the original image size using bicubic interpolation, then sum them over the heads, layers, and time steps:
\begin{equation}\label{eqn:summation}\vspace{-2mm}
\small
D_k^\mathbb{R}[x,y] := \sum_{i,j,\ell}  \tilde{F}^{(i)\downarrow}_{t_j,k,\ell}[x, y] + \tilde{F}^{(i)\uparrow}_{t_j,k,\ell}[x,y],
\end{equation}
where $k$ is the $k^\text{th}$ word and $\tilde{F}^{(i)\downarrow}_{t_j,k,\ell}[x,y]$ is shorthand for $F^{(i)\downarrow}_t[x, y, \ell, k]$, bicubically upscaled to fixed size $(w, h)$.\footnote{We show that aggregating across \textit{all} time steps and layers is indeed necessary in Section~\ref{appendix:obj_attr}.}
Since $D_k^\mathbb{R}$ is positive and scale normalized (summing normalized values preserves linear scale), we can visualize it as a soft heat map, with higher values having greater attribution.
To generate a hard, binary heat map (either a pixel is influenced or not), we can threshold $D_k^\mathbb{R}$ as\vspace{-1mm}
\begin{equation}\label{eqn:threshold}\small\vspace{-2mm}
    D_k^\mathbb{I_\tau}[x, y] := \mathbb{I}\left(D_k^\mathbb{R}[x, y] \geq \tau \max_{i,j} D_k^\mathbb{R}[i,j]\right),
\end{equation}
where $\mathbb{I}(\cdot)$ is the indicator function and $\tau \in [0, 1]$.
See Figure~\ref{fig:daam} for an illustration of DAAM.

\section{Attribution Analyses}

\subsection{Object Attribution}\label{sec:obj-attr}

Quantitative evaluation of our method is challenging, but we can attempt to draw upon existing annotated datasets and methods to see how well our method aligns.
A popular visuosemantic task is image segmentation, where areas~(i.e., segmentation masks) are given a semantically meaningful label, commonly nouns.
If DAAM is accurate, then our attention maps should arguably align with the image segmentation labels for these tasks---despite not having been trained to perform this task.


\begin{table}[t]\small
    \setlength{\tabcolsep}{1.5pt}
    \centering
    \begin{tabular}{rlcccc}
    \toprule[1pt]
    \multirow{2}{*}{\#} & \multirow{2}{*}{Method} & \multicolumn{2}{c}{COCO-Gen} & \multicolumn{2}{c}{Unreal-Gen}\\
    \cmidrule(lr){3-4} \cmidrule(lr){5-6}
    & & {\tiny mIoU$^{80}$} & {\tiny mIoU$^\infty$} & {\tiny mIoU$^{80}$} & {\tiny mIoU$^{\infty}$}\\
    \midrule
    \multicolumn{6}{c}{Supervised Methods}\\
    \midrule
    1 & Mask R-CNN (ResNet-101) & 82.9 & 32.1 & 76.4 & 31.2\\
    2 & QueryInst (ResNet-101-FPN) & 80.8 & 31.3 & 78.3 & 35.0\\
    3 & Mask2Former (Swin-S) & \textbf{84.0} & 32.5 & \textbf{80.0} & 36.7\\
    4 & CLIPSeg & 78.6 & \textbf{71.6} & 74.6 & \textbf{70.9}\\
    \midrule
    \multicolumn{6}{c}{Unsupervised Methods}\\
    \midrule
    5 & Whole image mask & 20.4 & 21.1 & 19.5 & 19.3 \\
    6 & PiCIE + H & 31.3 & 25.2 & 34.9 & 27.8 \\
    7 & STEGO (DINO ViT-B) & 35.8 & 53.6 & 42.9 & 54.5 \\
    8 & Our DAAM-0.3 & 64.7 & 59.1 & 59.1 & \textbf{58.9} \\
    9 & Our DAAM-0.4 & \textbf{64.8} & \textbf{60.7} & \textbf{60.8} & 58.3\\
    10 & Our DAAM-0.5 & 59.0 & 55.4 & 57.9 & 52.5\\
    \bottomrule[1pt]
    \end{tabular}
    \caption{MIoU of semantic segmentation methods on our synthesized datasets. Best in each section bolded.}
    \label{tab:results}
\end{table}

\parheader{Setup}
We ran the Stable Diffusion 2.0 base model using 30 inference steps per image with the DPM~\cite{lu2022dpm} solver---see the appendix section~\ref{appendix:obj_attr} for specifics.
We then synthesized one set of images using the validation set of the COCO image captions dataset~\cite{lin2014microsoft}, representing realistic prompts, and another set by randomly swapping nouns in the same set (holding the vocabulary fixed), representing unrealism.
The purpose of the second set was to see how well the model generalized to uncanny prompts, whose composition was unlikely to have been encountered at training time.
We named the two sets ``COCO-Gen'' and ``Unreal-Gen,'' each with 100 prompt--image pairs.
For ground truth, we extracted all countable nouns from the prompts, then hand-segmented each present noun in the image.

To compute binary DAAM segmentation masks, we used Eqn.~\ref{eqn:threshold} with thresholds $\tau \in \{0.3, 0.4, 0.5\}$, for each noun in the ground truth.
We refer to these methods as DAAM-$\langle\tau\rangle$, e.g., DAAM-0.3.
For supervised baselines, we evaluated semantic segmentation models trained explicitly on COCO, like Mask R-CNN~\cite{he2017mask} with a ResNet-101 backbone~\cite{he2016deep}, QueryInst~\cite{fang2021instances} with ResNet-101-FPN~\cite{lin2017feature}, and Mask2Former~\cite{cheng2022mask2former} with Swin-S~\cite{liu2021swin}, all implemented in MMDetection~\cite{chen2019mmdetection}, as well as the open-vocabulary CLIPSeg~\cite{luddecke2022image} trained on the PhraseCut dataset~\cite{wu2020phrasecut}.
We note that CLIPSeg's setup resembles ours because the image captions are assumed given as well.
However, their model is supervised since they additionally train their model on segmentation labels.
Our unsupervised baselines consisted of the state-of-the-art STEGO~\cite{hamilton2021unsupervised} and PiCIE + H~\cite{cho2021picie}.
As is standard~\cite{lin2014microsoft}, we evaluated all approaches using the mean intersection over union (mIoU) over the prediction--ground truth mask pairs.
We denote mIoU$^{80}$ when restricted to the 80 COCO classes that the supervised baselines were trained on (save for CLIPSeg) and mIoU$^\infty$ as the mIoU without the class restriction.

\parheader{Results}
We present results in Table~\ref{tab:results}.
The COCO-supervised models (rows 1--3) are constrained to COCO's 80 classes (e.g., ``cat,'' ``cake''), while DAAM (rows 5--7) is open vocabulary; thus, DAAM outperforms them by 22--28 points in mIoU$^\infty$ and underperforms by 20 points in mIoU$^{80}$.
CLIPSeg (row 4), an open-vocabulary model trained on semantic segmentation datasets, achieves the best of both worlds in mIoU$^{80}$ and mIoU$^\infty$, with the highest mIoU$^\infty$ overall and high mIoU$^{80}$.
However, its restriction to nouns precludes it from generalized segmentation (e.g., verbs).
DAAM largely outperforms both unsupervised baselines (rows 6--7) by a margin of 4.4--29 points (see rows 7--10), likely because we assume the prompts to be provided.
Similar findings hold on the unrealistic Unreal-Gen set, showing that DAAM is resilient to nonsensical texts, confirming that DAAM works when Stable Diffusion has to generalize in composition.

As for $\tau$, 0.4 works best on all splits, though it isn't too sensitive, varying by 3--6 points in mIoU.
We also show that all layers and time steps contribute to DAAM's segmentation quality, shown in Section~\ref{appendix:obj_attr}.
Overall, DAAM forms a strong baseline of 57.9--64.8 mIoU$^{80}$.
We conclude that it is empirically sane, which we further support for all parts of speech in the next section.


\subsection{Generalized Attribution}
\begin{figure}
    \centering
    \includegraphics[scale=0.315]{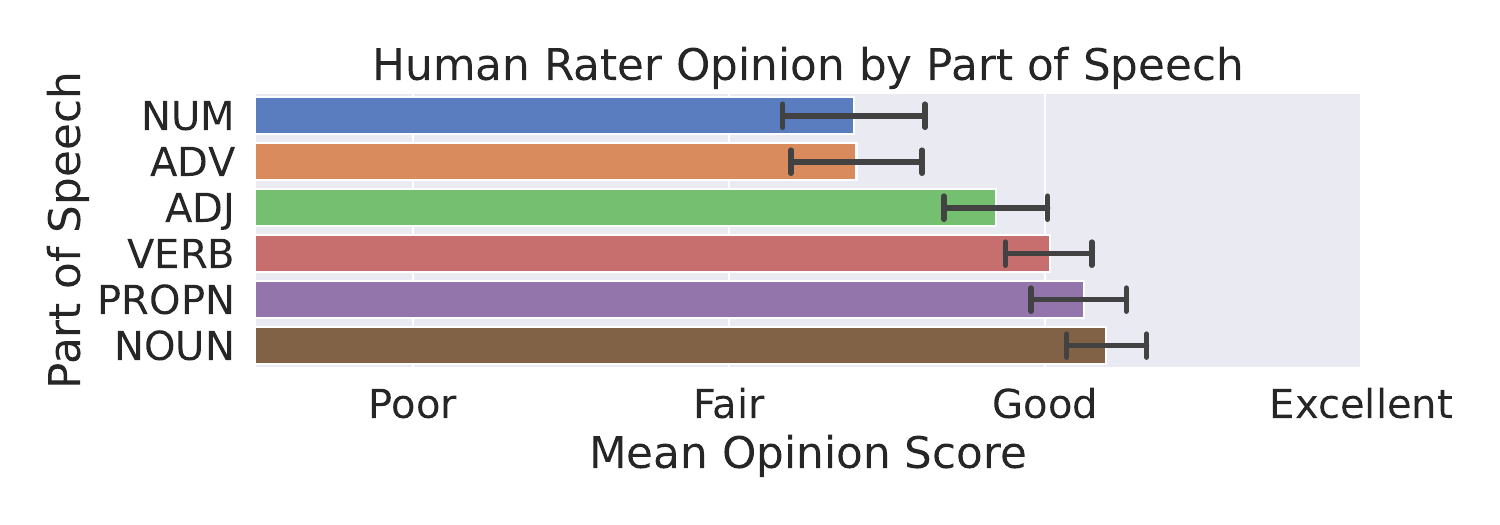}
    \includegraphics[scale=0.3]{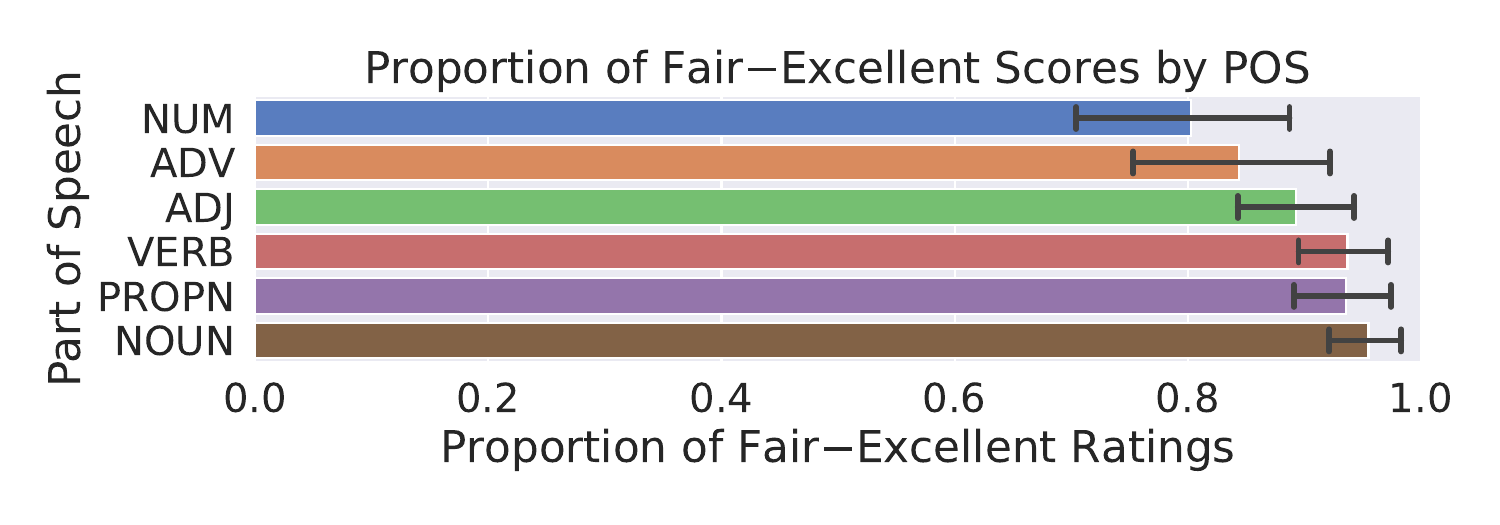}
    \caption{On the top, mean opinion scores grouped by part of speech, with 95\% confidence interval bars; on the bottom, proportion of fair--excellent scores, grouped by part-of-speech.}
    \label{fig:mos}
\end{figure}
We extend our veracity analyses beyond nouns to all parts of speech, such as adjectives and verbs, to show that DAAM is more generally applicable.
A high-quality, reliable analysis requires human annotation; hence, we ask human raters to evaluate the attribution quality of DAAM maps, using a five-point Likert scale.

This setup generalizes that of the last section because words in general are not visually separable, which prevents effective segmentation annotation.
For example, in the prompt ``people running,'' it is unclear where to visually segment ``running.''
Is it just the knees and feet of the runners, or is it also the swinging arms?
On the contrary, if annotators are instead given the proposed heat maps for ``running,'' they can make a judgement on how well the maps reflect the word.

\parheader{Setup}
To construct our word--image dataset, we first randomly sampled 200 words from each of the 14 most common part-of-speech tags in COCO, extracted with spaCy, for a total of 2,800 unique word--prompt pairs.
Next, we generated images alongside DAAM maps for all pairs, varying the random seed each time.
To gather human judgements, we built our annotation interface in Amazon MTurk, a crowdsourcing platform.
We presented the generated image, the heat map, and the prompt with the target word in red, beside a question asking expert workers to rate how well the highlighting reflects the word.
They then selected a rating among one of ``bad,'' ``poor,'' ``fair,'' ``good,'' and ``excellent'', as well as an option to declare the image itself as too poor or the word too abstract to interpret.
For quality control, we removed annotators failing attention tests.
For further robustness, we assigned three unique raters to each example.
We provide further details on the user interface and annotation process in the appendix section~\ref{appendix:gen_attr}.

\begin{figure}[t]
\centering
\includegraphics[scale=0.14]{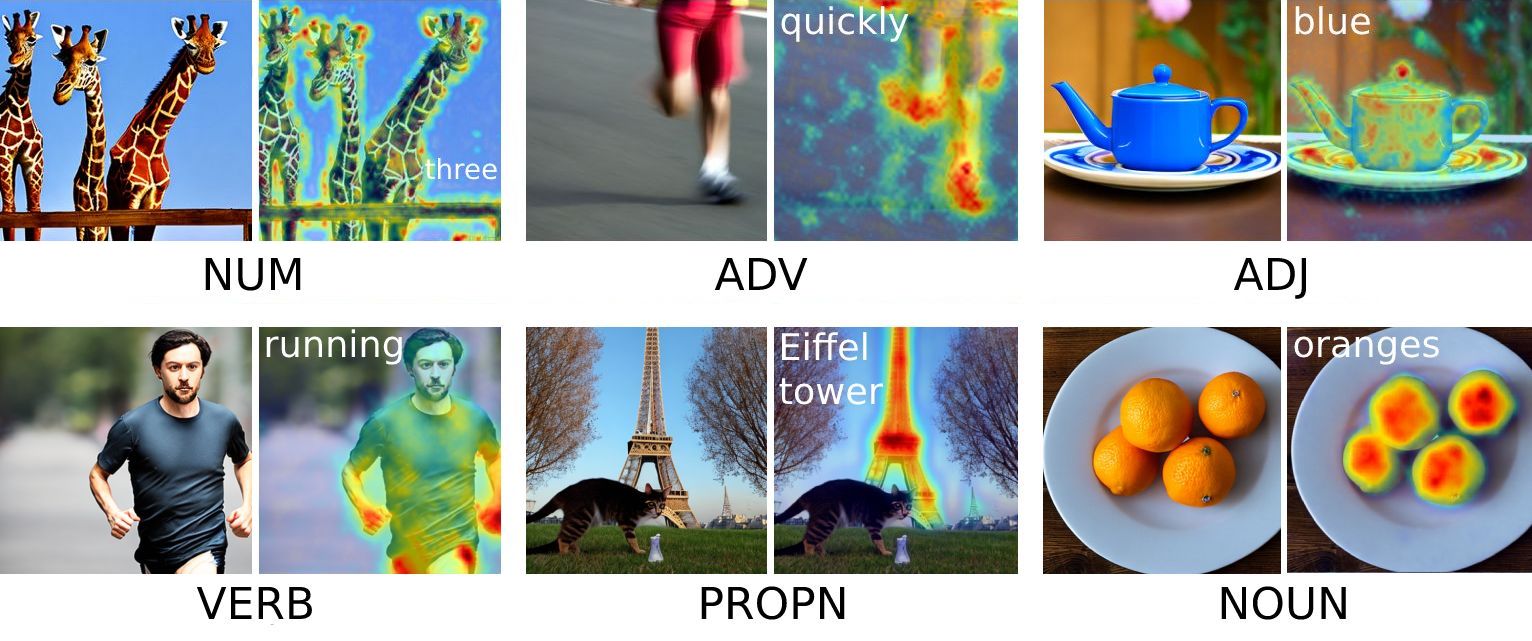}
\caption{Example generations and DAAM heat maps from COCO for each interpretable part-of-speech.}
\label{fig:mos-examples}
\end{figure}

\parheader{Results}
Our examples were judged by a total of fifty raters, none producing more than 18\% of the total number of annotations.
We filtered out all word--image pairs deemed too abstract (e.g., ``the''), when any one of the three assigned raters selected that option.
This resulted in six interpretable part-of-speech tags with enough judgements---see the appendix for detailed statistics.
To compute the final score of each word--image pair, we took the median of the three raters' opinions.

We plot our results in Figure~\ref{fig:mos}.
In the top subplot, we show that DAAM maps for adjectives, verbs, nouns, and proper nouns attain close to or slightly above ``good,'' whereas the ones for numerals and adverbs are closer to ``fair.''
This agrees with the generated examples in Figure~\ref{fig:mos-examples}, where numerals (see the giraffes' edges) and adverbs (feet and ground motion blur) are less intuitively highlighted than adjectives (blue part of teapot), verbs (fists and legs in running form), and nouns.
Nevertheless, the proportion of ratings falling between fair and excellent are above 80\% for numerals and adverbs and 90\% for the rest---see the bottom of Figure~\ref{fig:mos}.
We thus conclude that DAAM produces plausible maps for each interpretable part of speech.

One anticipated criticism is that different heat maps may explain the same word, making a qualitative comparison less meaningful.
In Figure~\ref{fig:mos-examples}, ``quickly'' could conceivably explain ``running'' too.
We concede to this, but our motivation is not to compare \textit{quality} but rather to demonstrate \textit{plausibility}.
Without these experiments, the DAAM maps for words like ``running'' and ``blue'' could very well have been meaningless blotches.

\section{Visuosyntactic Analysis}\label{sec:visuosyntactic-analysis}
\begin{table}[t]
\small
    \setlength{\tabcolsep}{3.5pt}
    \centering
    \begin{tabular}{rlccc|c}
    \toprule[1pt]
    \# & Relation & mIo\textbf{D} & mIo\textbf{H} & $\Delta$ & mIoU\\
    \midrule
    1 & Unrelated pairs & 65.1 & 66.1 & 1.0 & 47.5 \\
    2 & All head--dependent pairs & 62.3 & 62.0 & 0.3 & 43.4 \\
    \midrule
    3 & \texttt{compound} & 71.3 & 71.5 & 0.2 & 51.1 \\
    4 & \texttt{punct} & 68.2 & 70.0 & 1.8 & 49.5 \\
    5 & \texttt{nconj:and} & 58.0 & 56.1 & 1.9 & 38.2 \\
    6 & \texttt{det} & 54.8 & 52.2 & 2.6 & 35.0 \\
    7 & \texttt{case} & 51.7 & 58.1 & 6.4 & 36.9 \\
    8 & \texttt{acl} & \textbf{67.4} & 79.3 & \underline{12.} & 55.4 \\
    9 & \texttt{nsubj} & 76.4 & \textbf{63.9} & \underline{12.} & 52.2 \\
    10 & \texttt{amod} & \textbf{62.4} & 77.6 & \underline{15.} & 51.1 \\
    11 & \texttt{nmod:of} & 73.5 & \textbf{57.9} & \underline{16.} & 47.5 \\
    12 & \texttt{obj} & 75.6 & \textbf{46.3} & \underline{29.} & 55.4 \\
    \midrule
    14 & Coreferent word pairs & 84.8 & 77.4 & 7.4 & \textbf{66.6}\\
    \bottomrule[1pt]
    \end{tabular}
    \caption{Head--dependent DAAM map overlap statistics across the ten most common relations in COCO. Bolded are the dominant maps, where the absolute difference $\Delta$ between mIoD and mIoH exceeds 10 points. All bolded numbers are significant ($p < 0.01$).} 
    \label{tab:deprels-results}\vspace{-2.75mm}
\end{table}
Equipped with DAAM, we now study how syntax relates to generated pixels.
We characterize pairwise interactions between head--dependent DAAM maps, augmenting previous sections and helping to form hypotheses for further research.

\parheader{Setup}
We randomly sampled 1,000 prompts from COCO, performed dependency parsing with CoreNLP~\cite{manning2014stanford}, and generated an image for each prompt and DAAM maps for all words.
We constrained our examination to the top-10 most common relations, resulting in 8,000 head--dependent pairs.
Following Section~\ref{sec:obj-attr}, we then binarized the maps to quantify head--dependent interactions with set-based similarity statistics.
We computed three statistics between the DAAM map of the head and that of the dependent: first, the mean visual intersection area over the union (mIoU), i.e., $\scriptstyle \frac{|A \cap B|}{|A \cup B|}$; second, the mean intersection over the dependent (mIoD; $\scriptstyle \frac{|A \cap B|}{|A|}$); and third, the intersection over the head (mIoH; $\scriptstyle \frac{|A \cap B|}{|B|}$).
MIoU measures similarity, and the difference between mIoD and mIoH quantifies dominance.
If mIoD $>$ mIoH, then the head contains (dominates) the dependent more, and vice versa---see Appendix~\ref{appendix:syntactic} for a visual tutorial.

\parheader{Results}
\begin{figure}[t]
    \centering
    \includegraphics[scale=0.135]{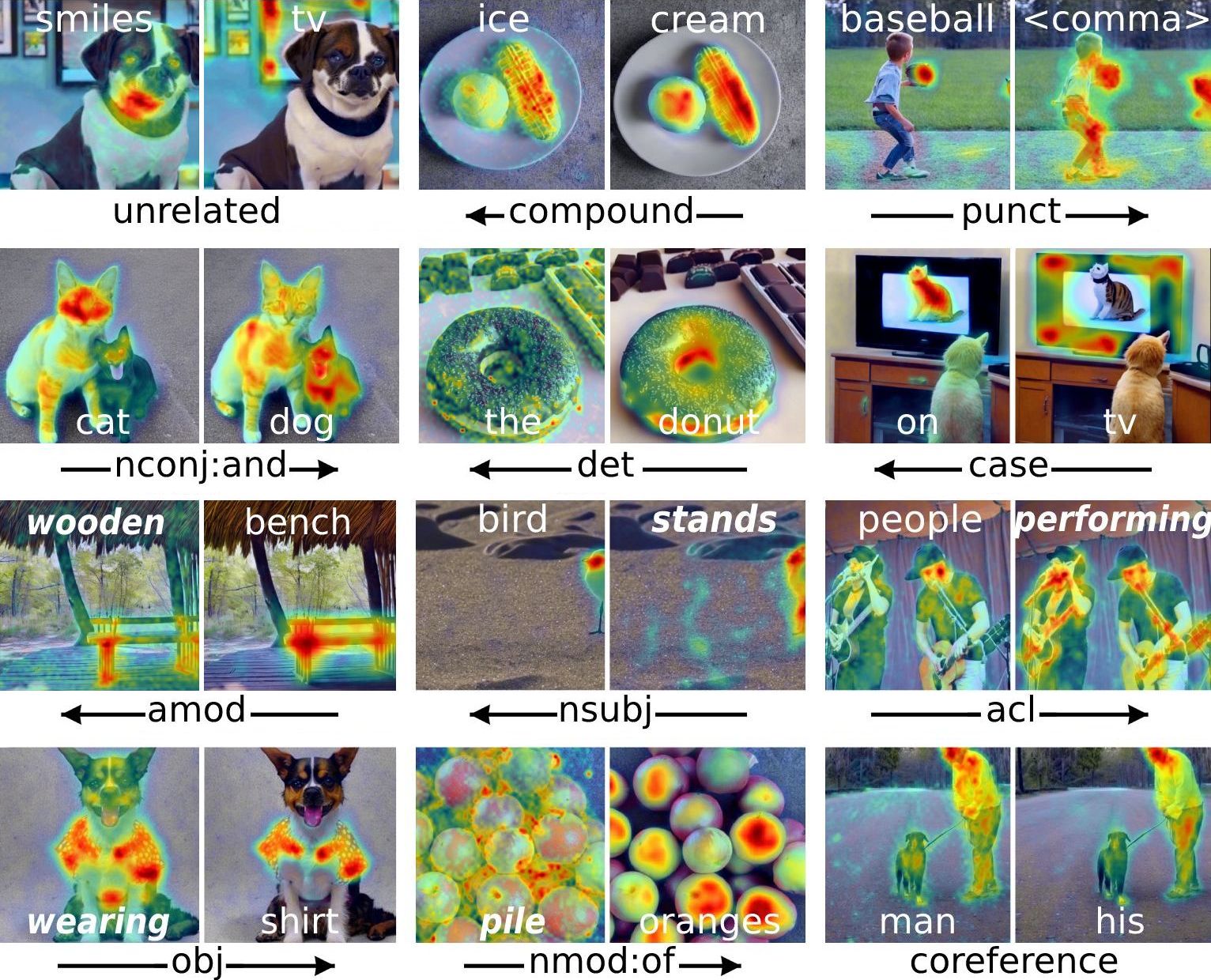}
    \caption{Twelve example pairs of DAAM maps, with the dominant word in bold, if present for the relation. Note that the visualization scale is normalized for each image since our purpose is to study the \textit{spatial locality} of attribution \textit{conditioned on the word}. For example, the absolute magnitude for the comma above is weak.}
    \label{fig:deprels-examples}
\end{figure}
We present our quantitative results in Table~\ref{tab:deprels-results} and examples in Figure~\ref{fig:deprels-examples}.
For baselines, we computed overlap statistics for unrelated pairs of words and all head--dependent pairs.
Unsurprisingly, both baselines show moderate similarity and no dominance (43--48 mIoU, $\Delta \leq 1$; rows 1--2).
For syntactic relations, we observe no dominance for noun compounds (row 3), which is expected since the two nouns complement one another (e.g., ``ice cream'').
Punctuation and articles (\texttt{punct}, \texttt{det}; rows 4 and 6) also lack dominance, possibly from having little semantic meaning and attending broadly across the image (Figure~\ref{fig:deprels-examples}, top right).
This resembles findings in \citet{kovaleva2019revealing}, who note BERT's~\cite{devlin2019bert} punctuation to attend widely.
For nouns connected with ``and'' (row 5), the maps overlap less (38.7 mIoU vs. 50+), likely due to visual separation (e.g., ``\textbf{cat} and \textbf{dog}'').
However, the overlap is still far above zero, which we attribute partially to feature entanglement, further explored in Section~\ref{sec:cohypo-entanglement}.

Starting at row 8, we arrive at pairs where one map dominates the other.
A group in core arguments arises (\texttt{nsubj}, \texttt{obj}), where the head word dominates the noun subject's or object's map (12--29-point $\Delta$), perhaps since verbs contextualize both the subject and the object in its surroundings---see the middle of and bottom left of Fig.~\ref{fig:deprels-examples}.
We observe another group in nominal dependents (\texttt{nmod:of}, \texttt{amod}, \texttt{acl}), where \texttt{nmod:of} mostly points to collective nouns (e.g., ``\textbf{pile} of \textbf{oranges}''), whose dominance is intuitive.
In contrast, adjectival modifiers (\texttt{amod}) behave counterintuitively, where descriptive adjectives (dependents) visually dominate the nouns they modify ($\Delta \approx 15$).
We instead expect objects to contain their attributes, but this is not the case.
We again ascribe this to entanglement, elucidated in Section~\ref{sec:adj-entanglement}.
Lastly, coreferent word pairs exhibit the highest overlap out of all relations (66.6 mIoU), indicating attention to the same referent.

\section{Visuosemantic Analyses}
\begin{figure}[t]
    \centering
    \includegraphics[scale=0.325,clip,trim={0 20 0 20}]{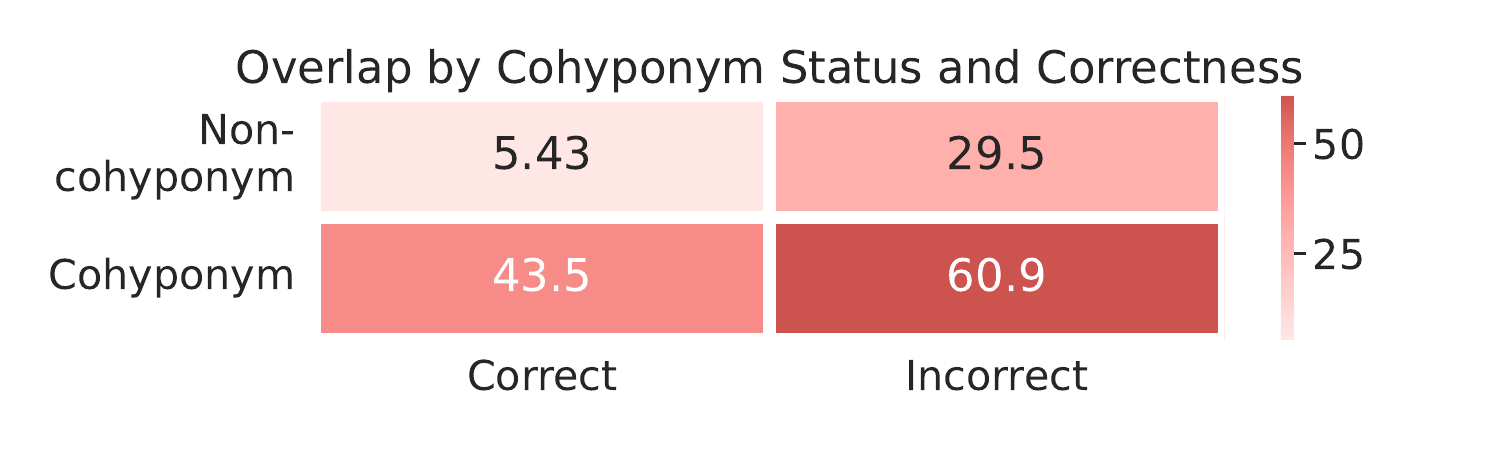}
    \centering
    \includegraphics[scale=0.325,clip,trim={0 20 0 0}]{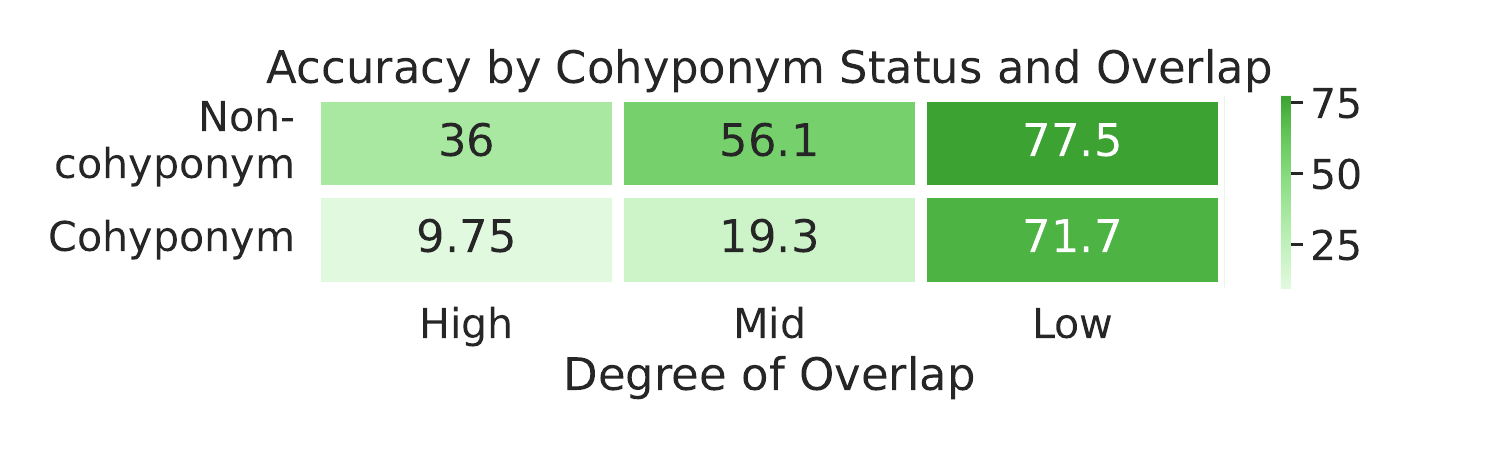}
    \caption{Above: DAAM map overlap in mean IoU, subdivided by cohyponym status and correctness; below:~generation accuracy, subdivided by cohyponym status and amount of overlap.}
    \label{fig:cohypo}
\end{figure}

\subsection{Cohyponym Entanglement}\label{sec:cohypo-entanglement}

To further study the large \texttt{nconj:and} overlap found in Section~\ref{sec:visuosyntactic-analysis}, we hypothesize that semantically similar words in a prompt have worse generation quality, where only one of the words is generated in the image, not all.

\parheader{Setup}
To test our hypothesis, we used WordNet~\cite{miller1995wordnet} to construct a hierarchical ontology expressing semantic fields over COCO's 80 visual objects, of which 28 have at least one other cohyponym across 16 distinct hypernyms (as listed in the appendix).
Next, we used the prompt template, ``a(n) \texttt{<noun>} and a(n) \texttt{<noun>},'' depicting two distinct things, to generate our dataset.
Using our ontology, we randomly sampled two cohyponyms 50\% of the time and two non-cohyponyms other times, producing 1,000 prompts from the template (e.g., ``a \textbf{giraffe} and a \textbf{zebra},'' ``a \textbf{cake} and a \textbf{bus}'').
We generated an image for each prompt, then asked three unique annotators per image to select which objects were present, given the 28 words.
We manually verified the image--label pairs, rejecting and republishing incorrect ones.
Finally, we marked the overall label for each image as the top two most commonly picked nouns, ties broken by submission order.
We considered generations correct if both words in the prompt were present in the image.
For more setup details, see the appendix.

\begin{figure}[t]
    \centering
    \includegraphics[scale=0.2]{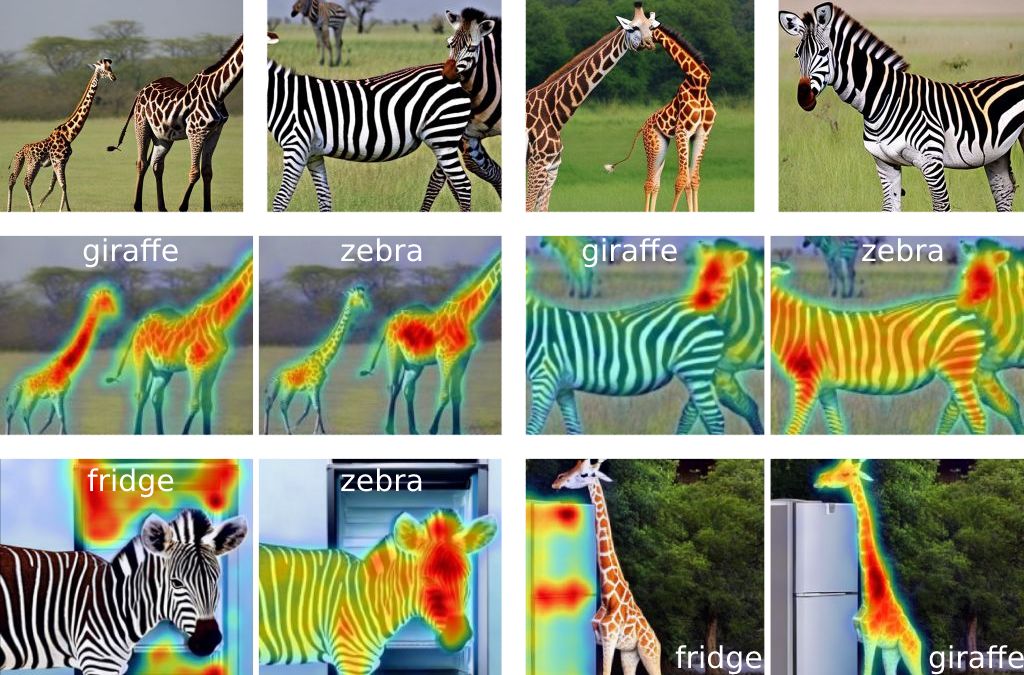}
    \caption{Rows starting from the top: generated images for cohyponyms ``a giraffe and a zebra,'' heat maps for the first two images, and heat maps for non-cohyponymic zebra--fridge and giraffe--fridge prompts.}
    \label{fig:cohypo-examples}
\end{figure}
\begin{figure}[t]
    \centering
    \includegraphics[scale=0.2]{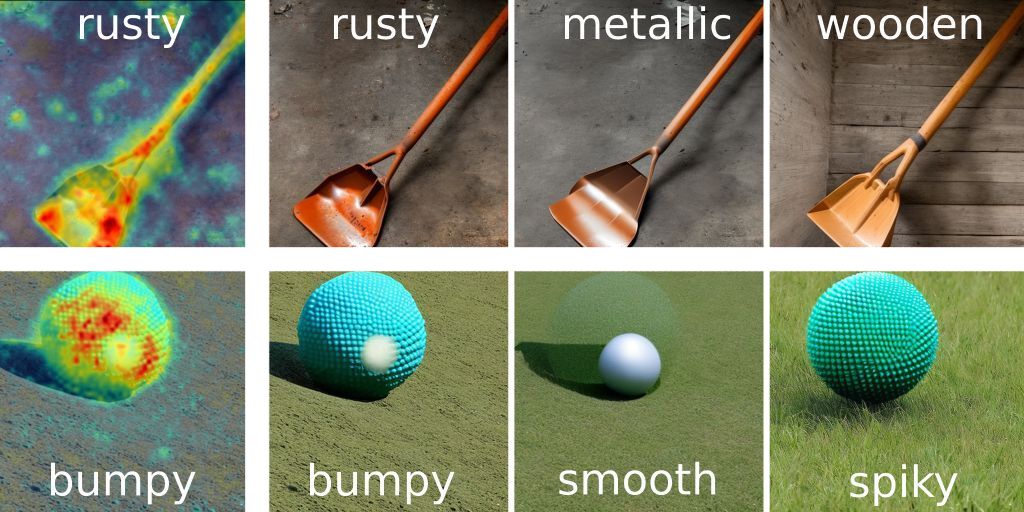}
    \caption{First row: a DAAM map for ``rusty'' and three generated images for ``a \texttt{<adj>} shovel sitting in a clean shed;'' second row: a map for ``bumpy'' and images for ``a \texttt{<adj>} ball rolling down a hill.''}
    \label{fig:adj-examples1}
\end{figure}

\parheader{Results}
Overall, the non-cohyponym set attains a generation accuracy of 61\% and the cohyponym set 52\%, statistically significant at the 99\% level according to the exact test, supporting our hypothesis.
To see if DAAM assists in explaining these effects, we compute binarized DAAM maps ($\tau=0.4$, the best value from Sec.~\ref{sec:obj-attr}) for both words and quantify the amount of overlap with IoU.
We find that the mIoU for cohyponyms and non-cohyponyms are 46.7 and 22.9, suggesting entangled attention and composition.
In the top of Figure~\ref{fig:cohypo}, we further group the mIoU by cohyponym status and correctness, finding that incorrectness and cohyponymy independently increase the overlap. 
In the bottom subplot, we show that the amount of overlap (mIoU) differentiates correctness, with the low, mid, and high cutoff points set at $\leq$ 0.4, 0.4--0.6, and $\geq$ 0.6, following statistics in Section~\ref{sec:visuosyntactic-analysis}.
We observe accuracy to be much better on pairs with low overlap (71.7--77.5\%) than those with high overlap (9.8--36\%).
We present some example generations and maps in Figure~\ref{fig:cohypo-examples}, which supports our results.

\subsection{Adjectival Entanglement}
\label{sec:adj-entanglement}
\begin{figure}[t]
    \centering
    \includegraphics[scale=0.2]{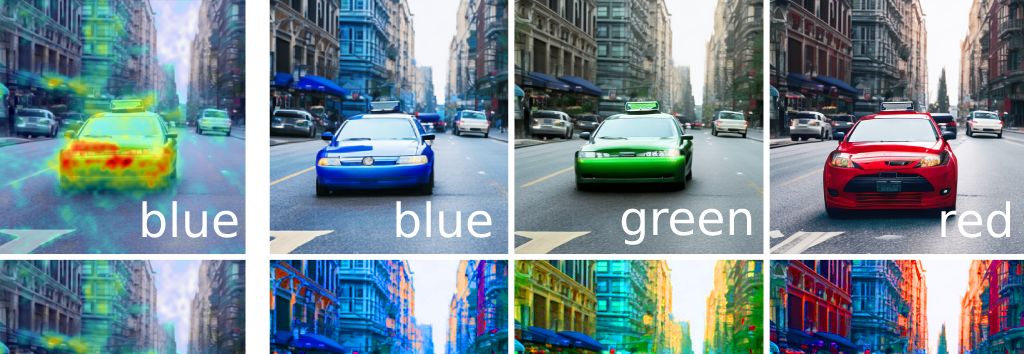}
    \caption{A DAAM map and generated images for ``a \texttt{<adj>} car driving down the streets,'' above images of the cropped background, saturated for visualization.}
    \label{fig:adj-examples2}
\end{figure}
We examine prompts where a noun's modifying adjective attends too broadly across the image.
We start with an initial seed prompt of the form, ``a \texttt{<adj>} \texttt{<noun>} \texttt{<verb phrase>},'' then vary the adjective to see how the image changes.
If there is no entanglement, then the background \textit{should not} gain attributes pertaining to that adjective.
To remove scene layout as a confounder, we fix all cross-attention maps to those of the seed prompt, which \citet{hertz2022prompt} show to equalize layout.

Our first case is, ``a \{rusty, metallic, wooden\} shovel sitting in a clean shed,'' ``rusty'' being the seed adjective.
As shown in Figure~\ref{fig:adj-examples1}, the DAAM map for ``rusty'' attends broadly, and the background for ``rusty'' is surely not clean.
When we change the adjective to ``metallic'' and ``wooden,'' the shed changes along with it, becoming grey and wooden, indicating entanglement.
Similar observations apply to our second case, ``a \{bumpy, smooth, spiky\} ball rolling down a hill,'' where ``bumpy'' produces rugged ground, ``smooth'' flatter ground, and ``spiky'' blades of grass.
In our third case, we study color adjectives using ``a \{blue, green, red\} car driving down the streets,'' presented in Figure~\ref{fig:adj-examples2}.
We discover the same phenomena, with the difference that these prompts lead to \textit{quantifiable} notions of adjectival entanglement.
For, say, ``green,'' we can conceivably measure the amount of additional green hue in the background, with the car cropped out---see bottom row.
A caveat is that entanglement is not necessarily unwanted; for instance, rusty shovels likely belong in rusted areas.
It strongly depends on the use case of the model.

\section{Related Work and Future Directions}
The primary area of this work is in understanding neural networks from the perspective of computational linguistics, with the goal of better informing future research.
A large body of relevant papers exists, where researchers apply textual perturbation~\cite{wallace2019allennlp}, attention visualization~\cite{vig2019bertviz, kovaleva2019revealing, shimaoka2016neural}, and information bottlenecks~\cite{jiang2020inserting} to relate important input tokens to the outputs of large language models.
Others explicitly test for linguistic constructs within models, such as \citeauthor{hendricks2021probing}'s (\citeyear{hendricks2021probing}) probing of vision transformers for verb understanding and \citeauthor{ilinykh2022attention}'s (\citeyear{ilinykh2022attention}) examination of visual grounding in image-to-text transformers.
Our distinction is that we carry out an attributive analysis in the space of generative diffusion models, as the pixel output relates to syntax and semantics.
As a future extension, we plan to assess the unsupervised parsing ability of Stable Diffusion with syntactic--geometric probes, similar to \citeauthor{hewitt2019structural}'s (\citeyear{hewitt2019structural}) work in BERT.

The intersection of text-to-image generation and natural language processing is certainly substantial.
In the context of enhancing diffusion models using prompt engineering, \citet{hertz2022prompt} cement cross-attention maps for the purpose of precision-editing generated images using text, and \citet{woolf} proposes negative prompts for removing undesirable, scene-wide attributes.
Related as well are works for generative adversarial networks, where \citet{karras2019style} and \citet{materzynska2022disentangling} disentangle various features such as style and spelling.
Along this vein, our work exposes more entanglement in cohyponyms and adjectives.
A future line of work is to disentangle such concepts and improve generative quality.

Last but not least are semantic segmentation works in computer vision.
Generally, researchers start with a backbone encoder, attach decoders, and then optimize the model in its entirety end-to-end on a segmentation dataset~\cite{cheng2022mask2former}, unless the context is unsupervised, in which case one uses contrastive objectives and clustering~\cite{cho2021picie, hamilton2021unsupervised}.
Toward this, DAAM could potentially provide encoder features in a segmentation pipeline, where its strong raw baseline numbers suggest the presence of valuable latent representations in Stable Diffusion.




\section{Conclusions}
In this paper, we study visuolinguistic phenomena in diffusion models by interpreting word--pixel cross-attention maps.
We prove the correctness of our attribution method, DAAM, through a quantitative semantic segmentation task and a qualitative generalized attribution study.
We apply DAAM to assess how syntactic relations translate to visual interactions, finding that certain maps of heads inappropriately subsume their dependents'.
We use these findings to form hypotheses about feature entanglement, showing that cohyponyms are jumbled and adjectives attend too broadly.

\section*{Acknowledgments}
Resources used in preparing this research were provided, in part, by the Province of Ontario, the Government of Canada through CIFAR, companies sponsoring the Vector Institute, and the HuggingFace team.
In particular, we would like to thank Aleksandra (Ola) Piktus, who helped us get a community grant for our public demonstration on HuggingFace spaces.

\bibliography{anthology}
\bibliographystyle{acl_natbib}

\appendix
\input{appendix}


\end{document}

%% file: appendix.tex
\section{Supplements for Attribution Analyses}

\subsection{Object Attribution}
\label{appendix:obj_attr}

\parheader{Generation setup}
For all generated images in the paper, we ran the Stable Diffusion 2.0 base model (512 by 512 pixels) with 30 inference steps, the default 7.5 classifier guidance score, and the state-of-the-art DPM solver.
We automatically filtered out all offensive images, against which the 2.0 model has both training-time and after-inference protection.
We also steered clear of offensive prompts, which were absent to start with in COCO.
Our computational environment consisted of PyTorch 1.11.0 and CUDA 11.4, running on Titan RTX and A6000 graphics cards.

\begin{figure}
\centering
\includegraphics[scale=0.25]{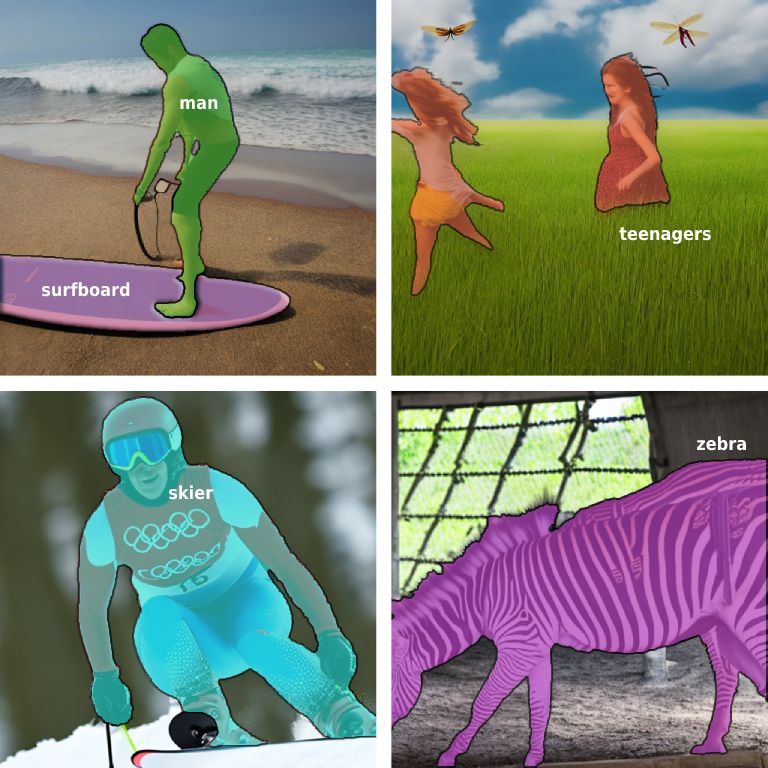}
\caption{Example ground-truth segmentation masks from four prompt--image pairs.}
\label{fig:segmentation-examples}
\end{figure}

\parheader{Segmentation process}
To draw the ground-truth segmentation masks, we used the object selection tool, the quick selection tool, and the brush from Adobe Photoshop CC 2022 to fill in a black mask for each area corresponding to a present noun.
We then exported each mask (without the background image) as a binary PNG mask and attached it to the relevant noun---see Figure~\ref{fig:segmentation-examples} for some examples.
Two trained annotators worked on the total set of 200 image--prompt pairs, with one completing 75 on each dataset and the other 25 on each.

\parheader{Layer and time step ablation}
\begin{figure}
    \centering
    \includegraphics[scale=0.45,clip,trim={7 0 7 0}]{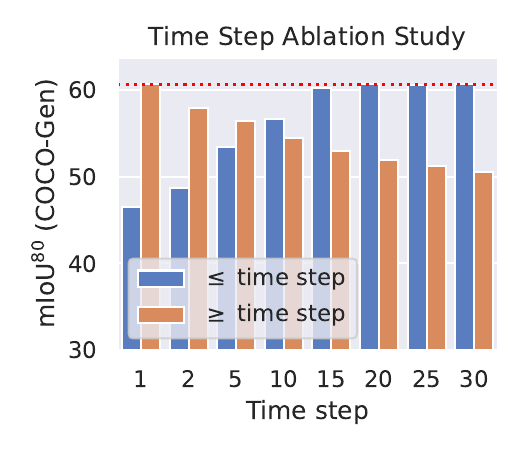}
    \includegraphics[scale=0.45,clip,trim={7 0 7 0}]{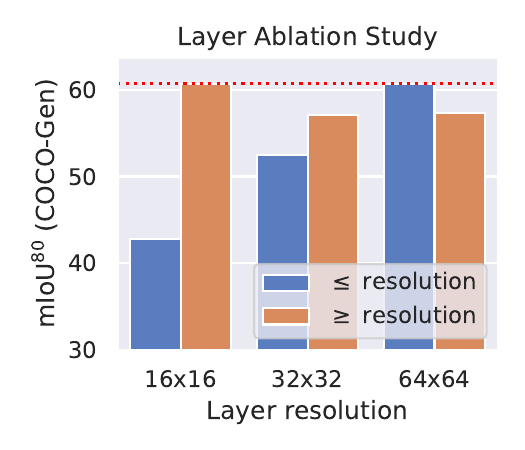}
    \caption{On the left, taking the first $n$ or last $n$ time steps; on the right, the equivalent for layer resolution.}
    \label{fig:ablation-studies}
\end{figure}
We conducted ablation studies to see if summing across \textit{all} time steps and layers, as in Eqn.~\ref{eqn:summation}, is necessary.
We searched both sides of the summation: for one study, we restricted DAAM to $j\leq j^*$, as $j^* = 1 \to T$; for its dual study, we constrained $j \geq j^*$.
We applied the same methods to layer resolution, i.e., $c^i$.
We present our results in Fig.~\ref{fig:ablation-studies}, which suggests that all time steps and layers contribute positively to segmentation quality.

\subsection{Generalized Attribution}
\label{appendix:gen_attr}

\begin{figure}
\centering
\includegraphics[scale=0.24]{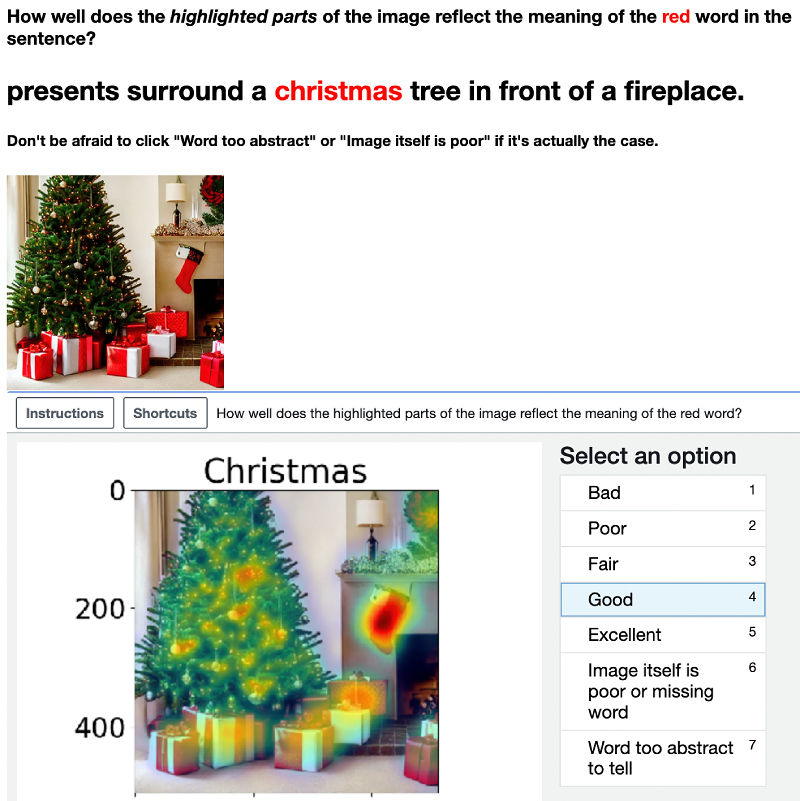}
\caption{Annotation UI for generalized attribution.}
\label{fig:annotation-ui1}
\centering
\includegraphics[scale=0.5]{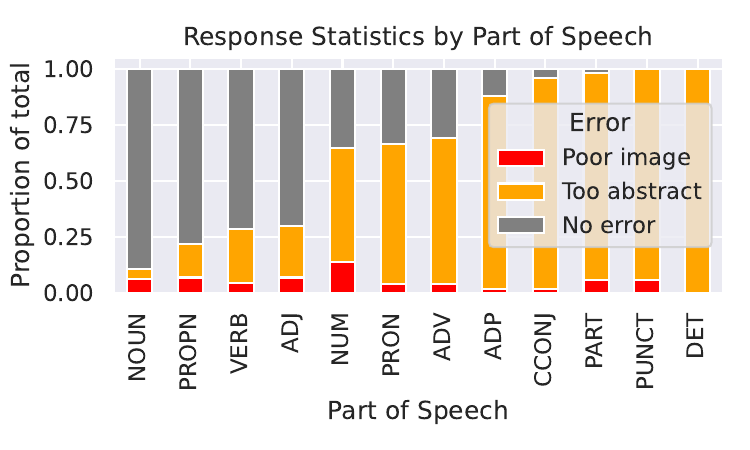}
\caption{Response statistics by part of speech.}
\label{fig:response-statistics}
\end{figure}

\parheader{Annotation process}
We designed our annotation user interfaces (UIs) for Amazon MTurk, a popular crowdsourcing platform, where we submitted a batch job requiring three unique annotators at the master level to complete each task.
We presented the UI pictured in Figure~\ref{fig:annotation-ui1}, asking them to rate the relevance of the red word to the highlighted area in the image.
If the image was too poor or if the word was missing, they could also choose options 6 and 7.

To filter out low-quality or inattentive annotators, we randomly asked workers to interpret punctuation, such as periods.
Since these tokens are self-evidently too abstract and missing in the image, we removed workers who didn't select one of those two options.
However, we found overall attention to be high, having a reject rate of less than 2\% of the tasks, consistent with \citeauthor{hauser2016attentive}'s (\citeyear{hauser2016attentive}) findings that MTurk users outperform subject pool participants.
We show response statistics in Figure~\ref{fig:response-statistics}, where adpositions, coordinating conjunctions, participles, punctuation, and articles have high non-interpretable rates.

\section{Supplements for Syntactic Analyses}
\label{appendix:syntactic}

\parheader{Measures of overlap}
We use three measures of overlap to characterize head--dependent map interactions: mean intersection over union (mIoU), intersection over the dependent (mIoD), and intersection over the head (mIoH).
When mIoU is high, the maps overlap greatly; when mIoD is high but mIoH is low, the head map occupies more of the dependent than the dependent does the head; when the opposite is true, the dependent occupies more.

\begin{figure}
\centering
\includegraphics[scale=0.7]{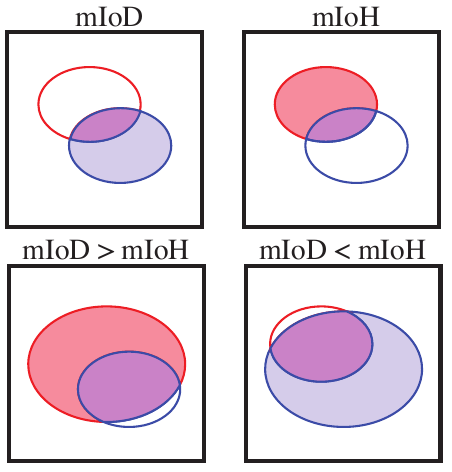}
\caption{Portrayals of mIoD, mIoH, and different forms of overlap.}
\label{fig:visual-overlap}
\end{figure}

Concretely, given a sequence of binarized DAAM map pairs $\{( D_{(i1)}^\mathbb{I_\tau}, D_{(i2)}^\mathbb{I_\tau})\}_{i=1}^n$, where $i1$ are \textbf{dependent} indices and $i2$ \textbf{head} indices, we compute mIoU as
\begin{equation}
\frac{1}{n}\sum_{i=1}^n \frac{\sum_{(x, y)} D_{(i1)}^\mathbb{I_\tau}[x, y] \wedge D_{(i2)}^\mathbb{I_\tau}[x, y]}{\sum_{(x, y)} D_{(i1)}^\mathbb{I_\tau}[x, y] \vee D_{(i2)}^\mathbb{I_\tau}[x, y]},
\end{equation}
where $\wedge$ is the logical-and operator, returning 1 if both sides are 1, 0 otherwise, and $\vee$ the logical-or operator, returning 1 if at least one operand is 1, and 0 otherwise.
Let the top part of the inner fraction be the intersection, or \textsc{Int} for short.
Define mIoD as
\begin{equation}
\frac{1}{n}\sum_{i=1}^n \frac{\textsc{Int}}{\sum_{(x, y)} D_{(i1)}^\mathbb{I_\tau}[x, y]},
\end{equation}
and mIoH as
\begin{equation}
\frac{1}{n}\sum_{i=1}^n \frac{\textsc{Int}}{\sum_{(x, y)} D_{(i2)}^\mathbb{I_\tau}[x, y]},
\end{equation}
We visually present our mIoD and mIoH statistics in Figure~\ref{fig:visual-overlap}.

\section{Supplements for Semantic Analyses}

\parheader{Semantic relation ontology}
We present our relation ontology below, continued on the next page:

\dirtree{%
.1 [ROOT].
.2 [BAG].
.3 backpack.
.3 handbag.
.3 suitcase.
.2 [FOOD].
.3 [BAKED GOODS].
.4 cake.
.4 donut.
.3 [DISH].
.4 hot dog.
.4 pizza.
.4 sandwich.
.3 [FRUIT].
.4 apple.
.4 banana.
.4 orange.
.2 [ELECTRICAL DEVICE].
.3 [APPLIANCE].
.4 oven.
.4 refrigerator.
.4 toaster.
.3 [MONITOR DEVICE].
.4 cell phone.
.4 laptop.
.4 tv.
.2 [FURNITURE].
.3 bench.
.3 chair.
.3 couch.
.2 [KITCHENWARE].
.3 [CUTLERY].
.4 fork.
.4 knife.
.4 spoon.
.3 [VESSEL].
.4 bowl.
.4 cup.
.2 [MAMMAL].
.3 [FARM ANIMAL].
.4 cow.
.4 horse.
.4 sheep.
.3 [PETS].
.4 cat.
.4 dog.
.3 [WILD ANIMAL].
.4 bear.
.4 elephant.
.4 giraffe.
.4 zebra.
.2 [SPORTS].
.3 skateboard.
.3 snowboard.
.3 surfboard.
.2 [VEHICLE].
.3 [AUTOMOBILE].
.4 bus.
.4 car.
.4 truck.
.3 [BIKE].
.4 bicycle.
.4 motorcycle.
}

\begin{figure}[t]
\includegraphics[scale=0.125]{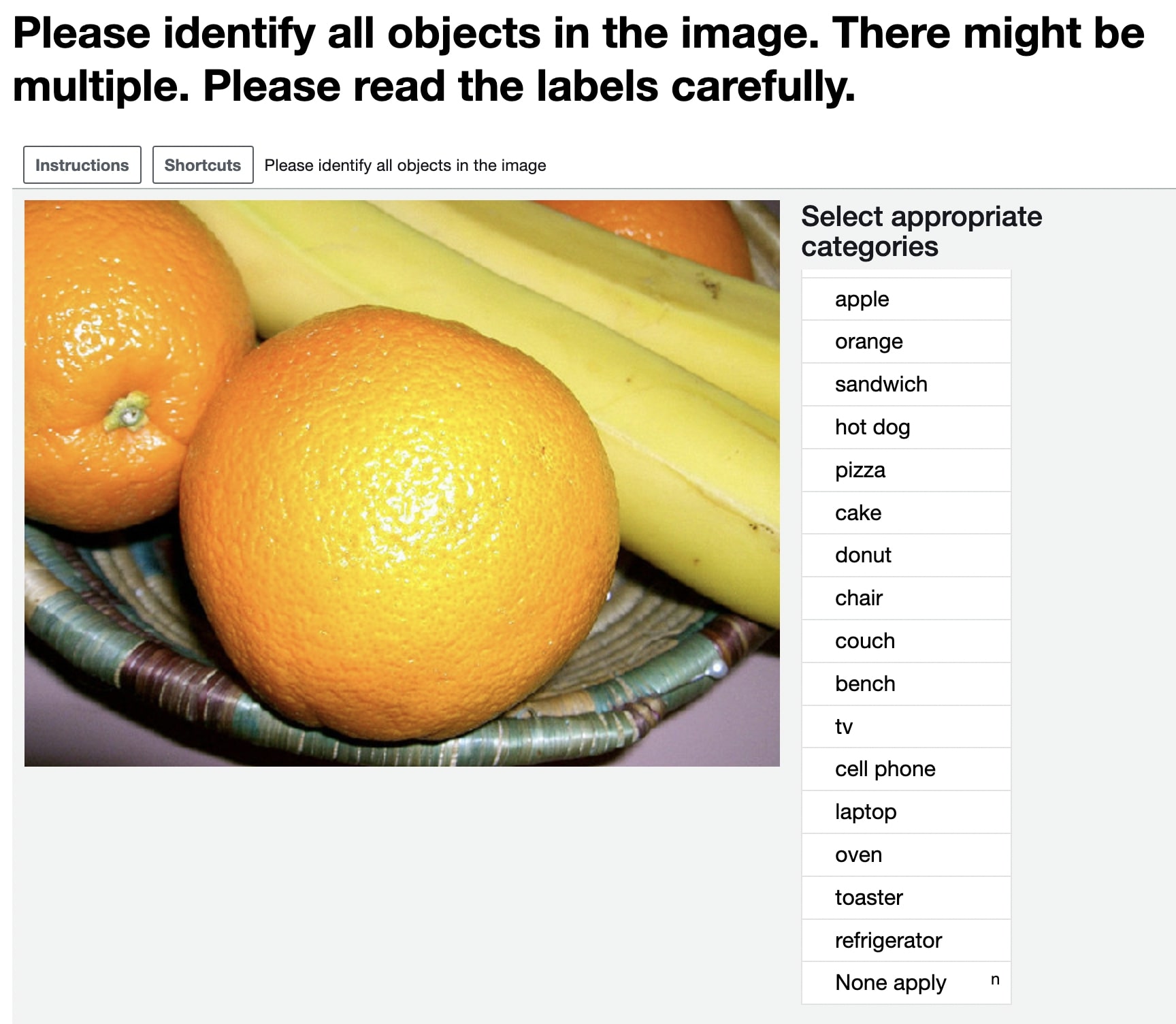}
\caption{The annotation UI for cohyponym entanglement, asking annotators to pick the present objects.}
\label{fig:annotation-ui2}
\end{figure}

\parheader{Cohyponym annotation process}
Similar to the generalized attribution annotation process, we designed our annotation UIs for Amazon MTurk.
We submitted a job requiring three unique annotators at the master level to complete each task.
We presented to them the UI shown in Figure~\ref{fig:annotation-ui2}.
We manually verified each response, removing workers whose quality was consistently poor.
This included workers who didn't include all objects generated.
Overall, the worker quality was exceptional, with a reject rate below 2\%.
Out of a pool of 30 workers, no single worker annotated more than 16\% of the examples.